\def\year{2022}\relax
\documentclass[letterpaper]{article} 
\usepackage{aaai22}  
\usepackage{times}  
\usepackage{helvet}  
\usepackage{courier}  
\usepackage[hyphens]{url}  
\usepackage{graphicx} 
\usepackage{natbib}  
\usepackage{caption} 
\DeclareCaptionStyle{ruled}{labelfont=normalfont,labelsep=colon,strut=off} 
\usepackage{algorithm}
\usepackage{algorithmic}
\usepackage{amssymb}
\usepackage{booktabs}
\usepackage{mathtools} 
\usepackage{multirow}

\def\I{{\bf I}}
\def\J{{\bf J}}

\usepackage{newfloat}
\usepackage{listings}
\floatstyle{ruled}
\newfloat{listing}{tb}{lst}{}
\floatname{listing}{Listing}
 \usepackage{hyperref} 

\title{Learning V1 Simple Cells with Vector Representation of Local Content and Matrix Representation of Local Motion}
\author{
    Ruiqi Gao,\textsuperscript{\rm 1}\thanks{The author is now a Research Scientist at Google Research, Brain team.}
    Jianwen Xie, \textsuperscript{\rm 2}
    Siyuan Huang, \textsuperscript{\rm 3}
    Yufan Ren,\textsuperscript{\rm 4}
    Song-Chun Zhu, \textsuperscript{\rm 1,3}
    Ying Nian Wu \textsuperscript{\rm 1}
}
\affiliations {
    \textsuperscript{\rm 1} UCLA, 
    \textsuperscript{\rm 2} Baidu Research, 
        \textsuperscript{\rm 3} Beijing Institute for General Artificial Intelligence (BIGAI), \\
    \textsuperscript{\rm 4} École Polytechnique Fédérale de Lausanne (EPFL)\\
    ruiqig@google.com, jianwen@ucla.edu, huangsiyuan@ucla.edu, yufan.ren@epfl.ch, sczhu@stat.ucla.edu, ywu@stat.ucla.edu
}

\usepackage{bibentry}

\begin{document}

\maketitle

\begin{abstract}
This paper proposes a representational model for image pairs such as consecutive video frames that are related by local pixel displacements, in the hope that the model may shed light on motion perception in primary visual cortex (V1). The model couples the following two components: (1) the vector representations of local contents of images and (2) the matrix representations of local pixel displacements caused by the relative motions between the agent and the objects in the 3D scene. When the image frame undergoes changes due to local pixel displacements, the vectors are multiplied by the matrices that represent the local displacements. Thus the vector representation is equivariant as it varies according to the local displacements. Our experiments show that our model can learn Gabor-like filter pairs of quadrature phases. The profiles of the learned filters match those of simple cells in Macaque V1. Moreover, we demonstrate that the model can learn to infer local motions in either a supervised or unsupervised manner. With such a simple model, we achieve competitive results on optical flow estimation.
\end{abstract}

\section{Introduction} \label{sect:1}

Our understanding of the primary visual cortex or V1 \cite{hubel1959receptive} is still very limited \cite{olshausen2005close}. In particular, mathematical and representational models for V1 are still in short supply. Two prominent examples of such models are sparse coding \cite{olshausen1997sparse} and independent component analysis (ICA)  \cite{bell1997independent}. Although such models may not provide detailed explanations at the level of neuronal dynamics, they help us understand the computational problems being solved by V1.

\begin{figure}[h]
\centering
	\includegraphics[height=.4\linewidth]{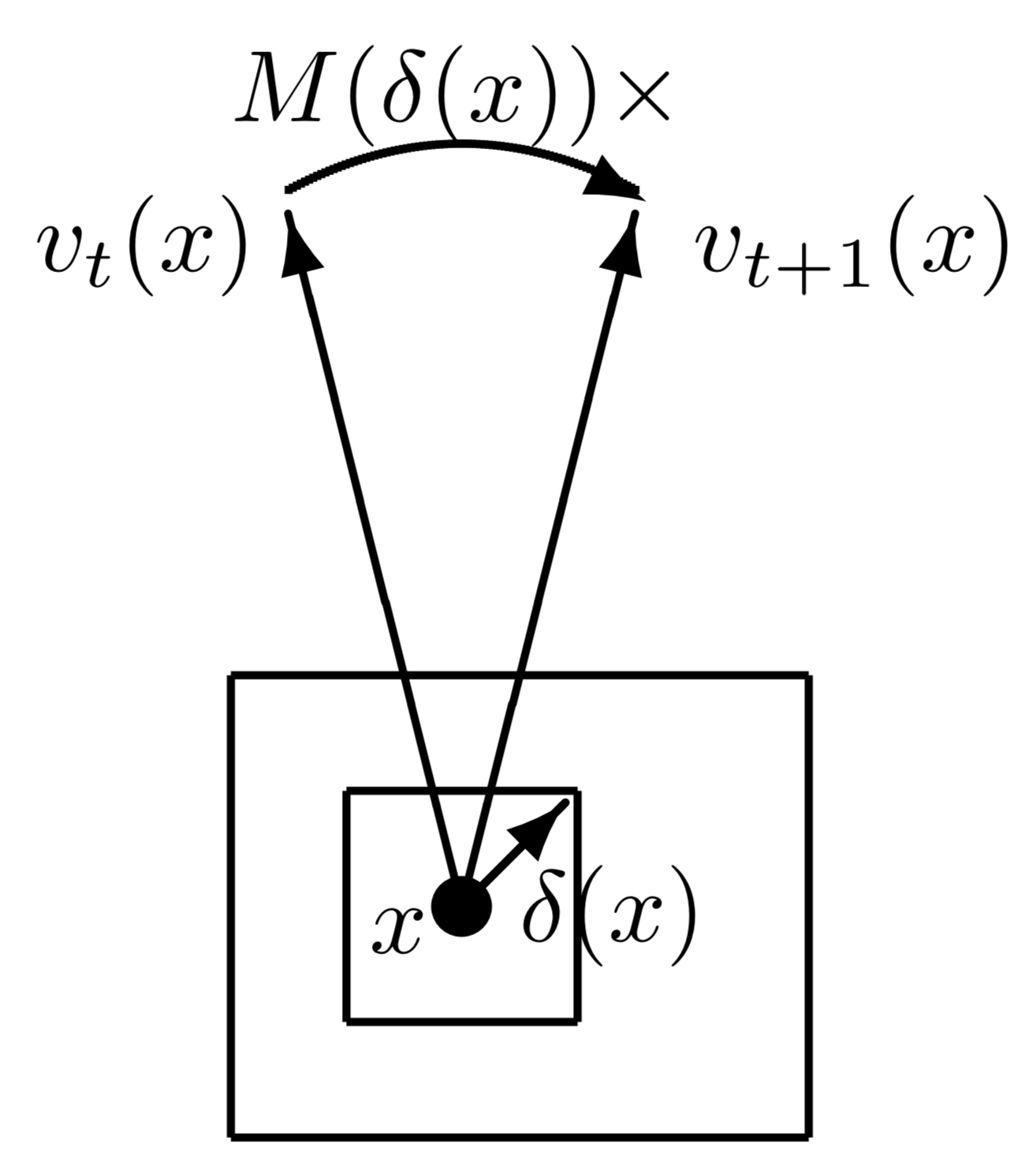}
	\caption{\small  Scheme of representation. The image is illustrated by the big rectangle. A pixel is illustrated by a dot. The local image content is illustrated by a small square around it. The displacement of the pixel is illustrated by a short arrow, which is within the small square. The vector representation of the local image content is represented by a long vector, which is equivariant because it rotates as the image undergoes deformation due to the pixel displacements. The rotation is realized by a matrix representation of the local motion. See Section \ref{sect:3} for the detailed notation.}
	\label{fig:diag}
	\end{figure}

In this paper, we propose a model of this sort. It is a representational model of natural image pairs that are related by local pixel displacements. The image pairs can be consecutive frames of a video sequence, where the local pixel displacements are caused by the relative motions between the agent and the objects in the 3D environment. Perceiving such local motions can be crucial for inferring ego-motion, object motions, and 3D depth information. 

As is the case with existing models, we expect our model to explain only partial aspects of V1, including: 
(1) The receptive fields of V1 simple cells resemble Gabor filters \cite{daugman1985uncertainty}. (2) Adjacent simple cells have quadrature phase relationship \cite{pollen1981phase,emerson1997quadrature}. (3) The V1 cells are capable of perceiving local motions. While existing models can all explain (1), our model can also account for (2) and (3) naturally.  Compared to models such as sparse coding and ICA, our model has a component that serves a direct purpose of perceiving local motions.

Our model consists of the following two components. 

(1) Vector representation of local image content.  The local content around each pixel is represented by a high dimensional vector.  Each unit in the vector is obtained by a linear filter. These local filters or wavelets are assumed to form a normalized wavelet tight frame, i.e., the image can be reconstructed from the vectors using the linear filters as the basis functions. 

(2) Matrix representation of local displacement. The change of the image from the current time frame to the next time frame is caused by the displacements of the pixels. Each possible displacement is represented by a matrix that operates on the vector. When the image changes according to the displacements, the vector at each pixel is multiplied by the matrix that represents the local displacement, in other words, the vector at each pixel is rotated by the matrix representation of the displacement of this pixel. Thus the vector representation is equivariant as it varies according to the local displacements. See Fig. \ref{fig:diag} for an illustration. 

We train the model on image pairs where in each pair, the second image is a deformed version of the first image, and the deformation is either known or inferred in training. We learn the encoding matrices for vector representation and the matrix representation of local displacements from the training data. 

Experiments show that our model learns V1-like units that can be well approximated by Gabor filters with quadrature phase relationship. The profiles of learned units match those of simples cells in Macaque V1. After learning the encoding matrices for vector representation and the matrix representations of the displacements, we can infer the displacement field using the learned model. Compared to popular optical flow estimation methods \cite{DFIB15,IMKDB17}, which use complex deep neural networks to predict the optical flows, our model is much simpler and is based on explicit vector and matrix representations. We demonstrate comparable performance to these methods in terms of the inference of displacement fields. 

In terms of biological interpretation, the vectors can be interpreted as activities of groups of neurons, and the matrices can be interpreted as synaptic connections. See Sections \ref{sect:b} and \ref{sect:s} for details.

\section{Contributions and related work} 

This paper proposes a simple representational model that couples the vector representations of local image contents and matrix representations of local pixel displacements. The model explains certain aspects of V1 simple cells such as Gabor-like receptive fields and quadrature phase relationship, and adds to our understanding of V1 motion perception in terms of representation learning. 

The following are two themes of related work. 

(1) V1 models. Most well known models for V1 are concerned with statistical properties of natural images or video sequences. Examples include sparse coding model \cite{olshausen1997sparse,lewicki1999probabilistic,olshausen2003learning}, independent component analysis (ICA) \cite{hyvarinen2004independent,bell1997independent,van1998independent}, slowness criterion  \cite{hyvarinen2003bubbles,wiskott2002slow}, and prediction \cite{singer2018sensory}. While these models are very compelling, they do not account for perceptual inference explicitly, which is the most important functionality of V1. On the other hand, our model is learned for the direct purpose of perceiving local motions caused by relative motion between the agent and the surrounding 3D environment. In fact, our model is complementary to the existing models for V1: similar to existing models, our work assumes a linear generative model for image frames, but our model adds a relational component with matrix representation that relates the consecutive image frames. 
Our model is also complementary to slowness criterion in that when the vectors are rotated by matrices, the norms of the vectors remain constant. 

(2) Matrix representation. In representation learning, it is a common practice to encode the signals or states as vectors.  However, it is a much less explored theme to represent the motions, actions or relations by matrices that act on the vectors. An early work in this theme is \cite{paccanaro2001learning}, which learns matrices to represent relations. More recently, \cite{jayaraman2015learning} learns equivariant representation with matrix representation of ego-motion. \cite{zhu2021learning} learns generative models of posed images based on invariant representation of 3D scene with matrix representation of ego-motion \cite{gao2018learning1, gao2021} learn vector representation of self-position and matrix representation of self-motion in a representational model of grid cells. Our work constitutes a new development along this theme. 

\section{Representational model} \label{sect:3}

\subsection{Vector representation} 

Let $\{\I(x), x \in D\}$ be an image observed at a certain instant, where $x = (x_1, x_2)  \in D$ is the 2D coordinates of pixel. $D$ is the image domain (e.g., $128 \times 128$). We represent the image $\I$ by vectors $\{v(x), x \in D_{-}\}$, where each $v(x)$ is a vector defined at pixel $x$, and $D_{-}$ may consist of a sub-sampled set of pixels in $D$. ${V} = \{v(x), x \in D_{-}\}$ forms a vector representation of the whole image. 

Conventionally, we assume the vector encoding is linear and convolutional. Specifically, let $\I[x]$ be a squared patch (e.g., $16 \times 16$) of $\I$ centered at $x$. We can flatten $\I[x]$ into a vector (e.g., 256 dimensional) and let 
\begin{eqnarray} 
   v(x) = W \I[x], \; x \in D_{-}, 
\end{eqnarray}
be the linear encoder, where $W$ is the encoding matrix that encodes $\I[x]$ into a vector $v(x)$. The rows of $W$ are the linear filters and can be displayed as local image patches of the same size as the image patch $\I[x]$. We can further write ${V} = {\bf W} {\bf I}$ if we treat ${\bf I}$ as a vector, and the rows of ${\bf W}$ are the translated versions of $W$.  

\subsection{Tight frame auto-encoder} 
We assume that ${\bf W}$ is an auto-encoding tight frame. Specifically, let $W(x)$ denote the translation of filter $W$ to pixel $x$ and zero-padding the pixels outside the filters, so that each row of $W(x)$ is of the same dimension as $\I$. We then assume
\begin{eqnarray}
\I = {\bf W}^{\top} {V} = \sum_{x \in D_{-}} W^{\top}(x)v(x),
\label{eq:tight}
\end{eqnarray} 
i.e., the linear filters for bottom-up encoding also serve as basis functions for top-down decoding. Both the encoder and decoder can be implemented by convolutional linear neural networks. 

The tight frame assumption can be justified by the fact that under that assumption, for two images ${\bf I}$ and ${\bf J}$, we have 
$ 
    \langle {\bf W} {\bf I}, {\bf W} {\bf J} \rangle = {\bf I}^\top {\bf W}^{\top} {\bf W} {\bf J} = \langle {\bf I}, {\bf J}\rangle, 
$
i.e., the vector representations preserve the inner product. As a result, we also have $\|{\bf W}\I\| = \|\I\|$ and $\|{\bf W} \J\| = \|\J\|$, so that the vector representations ${\bf W} {\bf I}$ and ${\bf W} {\bf J}$ preserve the angle between the images ${\bf I}$ and ${\bf J}$ and has the isometry property.  That is, when the image $\I$ changes from $\I_t$ to $\I_{t+1}$, its vector representation $V$ changes from $V_t$ to $V_{t+1}$, and the angle between $\I_t$ and  $\I_{t+1}$ is the same as the angle between $V_t$ and  $V_{t+1}$. 

In this paper, we assume a tight frame auto-encoder for computational convenience. A more principled treatment is to treat the decoder as a top-down generative model, and treat the encoder as approximate inference. Sparsity constraint can be imposed on the top-down decoder model. 

\subsection{Matrix representation}

Let $\I_t$ be the image at time frame $t$. Suppose the pixels of $\I_t$ undergo local displacements $(\delta(x), \forall x)$, where $\delta(x)$ is the displacement at pixel $x$. The image transforms from $\I_t$ to $\I_{t+1}$. We assume that $\delta(x)$ is within a squared range $\Delta$ (e.g., $[-6, 6] \times [-6, 6]$ pixels)  that is inside the range of the local image patch $\I_t[x]$ (e.g., $16 \times 16$ pixels). We assume that the displacement field $(\delta(x), \forall x)$ is locally smooth, i.e., pixels within each local image patch undergoes similar displacements. Let $v_t(x)$ and $v_{t+1}(x)$ be the vector representations of $\I_t[x]$ and $\I_{t+1}[x]$ respectively.  
The transformation from $\I_t[x]$ to $\I_{t+1}[x]$ is illustrated by the following diagram:
\begin{eqnarray}
\begin{array}[c]{ccc}
{v}_t(x) & \stackrel{ M(\delta(x)) \times }{\xrightarrow{\hspace*{1cm}}}& {v}_{t +1}(x) \\
&  & \\
W \uparrow& \uparrow  &\uparrow W\\
\I_t[x] &\stackrel{ \delta(x) }{\xrightarrow{\hspace*{1cm}}} & \I_{t+1}[x]
\end{array}  \label{eq:diagram}
\end{eqnarray}
Specifically, we assume that 
\begin{eqnarray} 
   {v}_{t+1}(x)  = M(\delta(x)) {v}_{t}(x),\; \forall x \in D_{-}.
   \label{eqn: motion}
\end{eqnarray} 
That is, when $\I$ changes from $\I_t$ to $\I_{t+1}$, ${v}(x)$ undergoes a linear transformation, driven by a matrix $M(\delta(x))$, which depends on the local displacement $\delta(x)$. Thus $v(x)$ is an equivariant representation as it varies according to $\delta(x)$. 

One motivation for modeling the transformation as a matrix representation operating on the vector representation of image patch comes from Fourier analysis.  Specifically, an image patch $\I[x]$ can be expressed by the Fourier decomposition $\I[x] = \sum_k c_k e^{i\langle \omega_k, x\rangle}$. Assuming all the pixels in the patch are shifted by a constant displacement $\delta(x)$, then the shifted image patch becomes $\I(x - \delta(x)) = \sum_k c_k e^{-i\langle \omega_k, \delta(x)\rangle} e^{i\langle \omega_k, x\rangle}$. The change from the complex number $c_k$ to $c_k e^{-i\langle \omega_k, dx\rangle}$ corresponds to rotating a 2D vector by a $2 \times 2$ matrix. However, we emphasize that our model does not assume Fourier basis or its localized version such as Gabor filters. The model figures it out with generic vector and matrix representations.

\paragraph{Parametrization.} \label{sect:p}
We consider two ways to parametrize the matrix representation of local displacement $M(\delta(x))$. First, we can discretize the displacement $\delta(x)$ into a finite set of possible values $\{\delta\}$, and we learn a separate $M(\delta)$ for each $\delta$. 
Second, We can learn a parametric version of $M(\delta)$ as the second order Taylor expansion of a matrix-valued function of $\delta= (\delta_1, \delta_2)$, 
\begin{eqnarray}
   M(\delta) = I + B_1 \delta_1 + B_2 \delta_2 + B_{11} \delta_1^2 + B_{22} \delta_2^2 + B_{12} \delta_1 \delta_2. \label{eq:taylor}
\end{eqnarray}
In the above expansion, $I$ is the identity matrix, and $B = (B_1, B_2, B_{11}, B_{22}, B_{12})$ are matrices of coefficients of the same dimensionality as $M(\delta)$. 

A more careful treatment is to treat $M(\delta)$ as forming a matrix Lie group, and write $M(\delta)$ as an exponential map of generator matrices that span the matrix Lie algebra. By assuming skew-symmetric generator matrices, $M(\delta)$ will be automatically a rotation matrix. The exponential map can be approximated by Taylor expansion similar to \ref{eq:taylor}. 

\paragraph{Local mixing.}
If $\delta(x)$ is large, $v_{t+1}(x)$ may contain information from adjacent image patches of $\I_{t}$ in addition to $\I_t[x]$. To address this problem, we can generalize the motion model (Eq. (\ref{eqn: motion})) to allow local mixing of encoded vectors. Specifically, let $\mathcal{S}$ be a local support centered at $0$. We assume that
\begin{eqnarray}
	{v}_{t+1}(x)  = \sum_{\text{d} x \in \mathcal{S}}  M(\delta(x), \text{d} x) {v}_{t}(x + \text{d} x)
	\label{eqn: local_mixing}
\end{eqnarray}
In the learning algorithm, we discretize $\text{d} x$ and learn a separate $M(\delta, \text{d} x)$ for each $\text{d} x$.

\subsection{Sub-vectors and disentangled rotations}
The vector $v(x)$ can be high-dimensional. For computational efficiency, we further divide $v(x)$ into $K$ sub-vectors, $v(x) = (v^{(k)}(x), k = 1, ..., K)$. Each sub-vector is obtained by an encoding sub-matrix $W^{(k)}$, i.e., 
\begin{eqnarray}
   v^{(k)}(x) = W^{(k)} \I[x], \; k = 1, ..., K, 
\end{eqnarray}
where $W^{(k)}$ consists of the rows of $W$ that correspond to $v^{(k)}$. In practice, we find that this assumption is necessary for the emergence of V1-like receptive field.

Correspondingly, the matrix representation 
\begin{eqnarray}
M(\delta) = \text{diag}(M^{(k)}(\delta), k = 1,..., K)
\end{eqnarray}
 is block diagonal. Each sub-vector $v^{(k)}(x)$ is transformed by its own sub-matrix $M^{(k)}(\delta)$: 
\begin{eqnarray}
 v^{(k)}_{t+1}(x) = M^{(k)}(\delta) v^{(k)}_t(x), \; k = 1, ..., K.
\end{eqnarray}
  The linear transformations of the sub-vectors $v^{(k)}(x)$ can be considered as rotations.  $v(x)$ is like a multi-arm clock, with each arm $v^{(k)}(x)$ rotated by $M^{(k)}(\delta(x))$. The rotations of $v^{(k)}(x)$ for different $k$ and $x$ are disentangled, meaning that the rotation of a sub-vector does not depend on other sub-vectors. 
  
 The assumption of sub-vectors and block-diagonal matrices is necessary for learning Gabor-like filters.  For ${v}_{t+1}(x)  = M(\delta(x)) {v}_{t}(x)$, it is equivalent to $\tilde{v}_{t+1}(x)  = \tilde{M}(\delta(x)) \tilde{v}_{t}(x)$ if we let $\tilde{v}_{t}(x) = P v_t(x)$, $\tilde{v}_{t+1}(x) = P v_{t+1}(x)$, and $\tilde{M}(\delta(x)) = P M(\delta(x)) P^{-1}$, for an invertible $P$. Assuming block-diagonal matrices helps eliminates such ambiguity. More formally, a matrix representation is irreducible if it cannot be further diagonalized into smaller block matrices. Assuming block-diagonal matrices helps to make the matrix representation close to irreducible.

\section{Learning and inference} 

 \subsection{Supervised learning} 
 \label{sect: loss}
 The input data consist of the triplets $(\I_t, (\delta(x), x \in D_{-}), \I_{t+1})$, where $(\delta(x), x \in D_{-})$ is the given displacement field. 
The unknown parameters to learn consist of matrices $(W, M(\delta), \delta \in \Delta)$, where $\Delta$ is the range of $\delta$. In the case of parametric $M$, we learn the $B$ matrices in the second order Taylor expansion (Eq. (\ref{eq:taylor})). We assume that there are $K$ sub-vectors so that $M$ or $B$ are block-diagonal matrices. We learn the model by optimizing a loss function defined as a weighted sum of two loss terms, based on the linear transformation model (Eq. (\ref{eqn: motion})) and tight frame assumption (Eq. (\ref{eq:tight})) respectively:\\
  (1) Linear transformation loss 
 \begin{eqnarray} 
    L_1 =   \sum_{k=1}^{K} \sum_{x \in D_{-}} \left\| W^{(k)}\I_{t+1}[x]  - M^{(k)}(\delta(x)) W^{(k)} \I_{t}[x] \right\|^2.
      \label{eqn: vector_loss}
\end{eqnarray} 
     For local mixing generalization, we substitute $M^{(k)}(\delta(x))$ by $\sum_{\text{d} x \in \mathcal{S}}  M^{(k)}(\delta(x), \text{d} x)$.  \vspace{0.8mm}\\
(2) Tight frame auto-encoder loss
     \begin{eqnarray} 
    L_{2} = \sum_{s \in {t, t+1}}   \left\| \I_{s} -  {\bf W}^\top {\bf W}\I_s\right\|^2. 
      \label{eqn: tight_frame_loss}
     \end{eqnarray}

\subsection{Inference of motion}
\label{sect: inference}
After learning $(W, M(\delta), \delta \in \Delta)$, given a testing pair  $(\I_t, \I_{t+1})$, we can infer the pixel displacement field $(\delta(x), x \in D_{-})$ by minimizing the linear transformation loss: $\delta(x) = \arg\max_{\delta \in \Delta} L_{1, x}(\delta)$, where 
\begin{equation}
\resizebox{.91\linewidth}{!}{$
  L_{1, x}(\delta) = \sum_{k=1}^{K} \left\| W^{(k)}\I_{t+1}[x]  - M^{(k)}(\delta) W^{(k)} \I_{t}[x] \right\|^2. $}\label{eq:infer} 
\end{equation}  
   This algorithm is efficient in nature as it can be parallelized for all $x \in D_{-}$ and for all $\delta \in \Delta$.  
   
If we use a parametric version of $(M(\delta), \delta \in \Delta)$ (Eq. (\ref{eq:taylor})), we can minimize $\sum_x L_{1, x}(\delta)$ using gradient descent with $\delta$ initialized from random small values. To encourage the smoothness of the inferred displacement field, we add a penalty term $\|\triangledown \delta(x)\|^2$ for this setting.

\subsection{Unsupervised learning}
We can easily adapt the learning of the model to an unsupervised manner, without knowing the pixel displacement field $(\delta(x), x \in D_{-})$. Specifically, we can iterate the following two steps: (1) update model parameters by loss functions defined in Section \ref{sect: loss}; (2) infer the displacement field as described in \ref{sect: inference}. To eliminate the ambiguity of $M(\delta)$ with respect to $\delta$, we add a regularization term $\left\| \I_{t+1} - {\rm warp}(\I_t, \delta) \right\|^2$ in the inference step, where ${\rm warp}(\cdot, \cdot)$ is a differentiable warping function, and we use the parametric version of $(M(\delta), \delta \in \Delta)$. To summarize, we infer the displacement field $(\delta(x), x \in D_{-})$ by minimizing:
\begin{eqnarray}
	\sum\nolimits_{x} L_{1, x}(\delta) + \|\triangledown \delta\|^2 + \left\| \I_{t+1} - {\rm warp}(\I_t, \delta) \right\|^2.
\end{eqnarray}
In practice, for each image pair at each iteration, we start the inference by running gradient descent on the inferred displacement field from the previous iteration.

\section{Discussions about model}

\subsection{Biological interpretations of cells and synaptic connections} \label{sect:b}

The learned $(W, M(\delta)), \delta)$ can be interpreted as synaptic connections. Specifically, for each block $k$, $W^{(k)}$ corresponds to one set of connection weights. Suppose $\delta \in \Delta$ is discretized, then for each $\delta$, $M^{(k)}(\delta)$ corresponds to one set of connection weights, and $(M^{(k)}(\delta), \delta \in \Delta)$ corresponds to multiple sets of connection weights. For motion inference in a biological system, after computing $v_{t, x}^{(k)} = W^{(k)} \I_{t}[x]$,  $M^{(k)}(\delta) v_{t, x}^{(k)}$ can be computed simultaneously for every $\delta \in \Delta$. Then $\delta(x)$ is inferred by max pooling according to Eq. (\ref{eq:infer}). 

$v_{t, x}^{(k)}$ can be interpreted as activities of simple cells, and $\|v_{t, x}^{(k)}\|^2$ can be interpreted as activity of a complex cell. If $M^{(k)}(\delta)$ is close to a rotation matrix, then we have norm stability so that $\|v_{t, x}^{(k)}\| \approx \|v_{t+1, x}^{(k)}\|$, which is closely related to the slowness property \cite{hyvarinen2003bubbles,wiskott2002slow}. 

\subsection{Spatiotemporal filters and recurrent implementation} \label{sect:s}

If we enforce norm stability or the orthogonality of $M^{(k)}(\delta)$, then minimizing $\|v_{t+1, x} - M(\delta) v_{t, x}\|^2$ over $\delta \in \Delta$ is equivalent to maximizing $\langle v_{t+1, x}, M(\delta) v_{t, x}\rangle$, which in turn is equivalent to maximizing $\|v_{t+1, x} + M(\delta) v_{t, x}\|^2$ so that $v_{t+1, x}$ and $M(\delta) v_{t, x}$ are aligned. This alignment criterion can be conveniently generalized to multiple consecutive frames, so that we can estimate the velocity at $x$ by maximizing the $m$-step alignment score $\|u\|^2$, where 
\begin{eqnarray}
u = \sum_{i = 0}^{m} M(\delta)^{m-i} v_{t+i, x} = \sum_{i=0}^{m} M(\delta)^{m-i} W \I_{t+i}[x]
\end{eqnarray} 
consists of responses of spatiotemporal filters or ``animated'' filters $(M(\delta)^{m-i} W, i = 0, ..., m)$, and $\|u\|^2$ corresponds to the energy of motion $\delta$ in the motion energy model \cite{adelson1985spatiotemporal} for direction selective cells. Thus our model is connected with the motion energy model. Moreover, our model enables a recurrent network for computing $u$ by $u_{i} = v_{t+i, x} + M(\delta) u_{i-1}$ for $i = 0, ..., m$, with $u_{-1} = 0$, and $u = u_m$. This recurrent implementation is much more efficient and biologically plausible than the plain implementation of spatiotemporal filtering which requires memorizing all the $\I_{t+i}$ for $i = 0, ..., m$. See \cite{pachitariu2017visual} for a discussion of biological plausibility of recurrent implementation of spatiotemporal filtering in general. 

The spatiotemporal filters can also serve as spatiotemporal basis functions for the top-down decoder model. 

\section{Experiments} 

The code, data and more results can be found at \url{http://www.stat.ucla.edu/~ruiqigao/v1/main.html}

We learn our model $(W, M(\delta), \delta \in \Delta)$ from image pairs $(\I_t, \I_{t+1})$ with its displacement field $(\delta(x))$ known or unknown. The number of sub-vectors $K = 40$, and the number of units in each sub-vector $v^{(k)}(x)$ is 2.  We use Adam \cite{kingma2014adam} optimizer for updating the model. To demonstrate the efficacy of the proposed model, we conduct experiments on two new synthetic datasets (V1Deform and V1FlyingObjects) and two public datasets (MPI-Sintel and MUG Facial Expression). The motivation to generate the synthetic datasets is that we find existing datasets such as Flying Chairs~\cite{DFIB15}, FlyingThings3D~\cite{mayer2016large}, and KITTI flow \cite{Geiger2012CVPR} contain image pairs with fairly large motions, which are unlikely consecutive frames perceived by V1. Thus we generate two synthetic datasets: V1Deform and V1FlyingObjects, which contains image pairs with only small local displacements and therefore better serve our purpose of studying motion perception in V1. See Fig. \ref{fig: infer_non_para1} for some examples from the synthetic datasets. 



\subsection{Datasets}
In this subsection, we elaborate the generation process of the two new synthetic datasets, and introduce the public datasets we use in this work. 

{\bf V1Deform.} For this dataset, we consider random smooth deformations for natural images. Specifically, We obtain the training data by collecting static images for $(\I_t)$ and simulate the displacement field $(\delta(x))$. The simulated displacement field is then used to transform $\I_t$ to obtain $\I_{t+1}$. We retrieve natural images as $\I_t$ from MIT places365 dataset \cite{zhou2016places}. The images are scaled to 128 $\times$ 128.  We sub-sample the pixels of images into a $m \times m$ grid ($m = 4$ in the experiments), and randomly generate displacements on the grid points, which serve as the control points for deformation. Then $\delta(x)$ for $x \in D$ can be obtained by spline interpolation of the displacements on the control points. We get $\I_{t+1}$ by warping $\I_t$ using $\delta(x)$ \cite{jaderberg2015spatial}. When generating a displacement $\delta = (\delta_1, \delta_2)$, both $\delta_1$ and $\delta_2$ are randomly sampled from a range of $[-6, +6]$. We synthesize $20,000$ pairs for training and $3,000$ pairs for testing. 
 
{\bf V1FlyingObjects.} For this dataset, we consider separating the displacement field into motions of the background and foreground, to jointly simulate the self-motion of the agent and the motion of the objects in the natural 3D scenes. To this end, we apply affine transformations to background images collected from MIT places365 \cite{zhou2016places} and foreground objects from a public 2D object dataset COIL-100 \cite{nene1996columbia}. The background images are scaled to $128 \times 128$, and the foreground images are randomly rescaled. To generate motion, we randomly sample affine parameters of translation, rotation, and scaling for both the foreground and background images. The motions of the foreground objects are relative to the background images, which can be explained as the relative motion between the moving object and agent. We tune the distribution of the affine parameters to keep the range of the displacement fields within $[-6, +6]$, which is consistent with the V1Deform dataset. Together with the mask of the foreground object and the sampled transformation parameters, we render the image pair $(\I_t, \I_{t+1})$ and its displacement field $(\delta(x))$ for each pair of the background image and foreground image. 

For the foreground objects, we obtain t he estimated masks from \cite{tev}, resulting in 96 objects with $72$ views per object available. We generate $14,411$ synthetic image pairs with their corresponding displacement fields and further split $12,411$ pairs for training and $2,000$ pairs for testing. Compared with previous optical flow dataset like Flying Chairs~\cite{DFIB15} and scene flow dataset like FlyingThings3D~\cite{mayer2016large}, the proposed V1FlyingObjects dataset has various foreground objects with more realistic texture and smoother displacement fields, which simulates more realistic environments. 

We shall release the two synthetic datasets, which are suitable for studying local motions and perceptions. Besides, we also use two public datasets:

{\bf MPI-Sintel.} MPI-Sintel \cite{Butler:ECCV:2012,Wulff:ECCVws:2012} is a public dataset designed for the evaluation of optical flow derived from rendered artificial scenes, with special attention to realistic image properties. Since MPI-Sintel is relatively small, which contains around a thousand image pairs, we use it only for testing the learned models in the inference of the displacement field. We use the final version of MPI-Sintel and resize each frame into size $128 \times 128$. We select frame pairs whose motions are within the range of $[-6, +6]$, resulting in $384$ frame pairs in total. 

{\bf MUG Facial Expression.} MUG Facial Expression dataset \cite{aifanti2010mug} records natural facial expression videos of $86$ subjects sitting in front of one camera. This dataset has no ground truth of the displacement field, which we use for unsupervised learning. $200$ videos with $30$ frames are randomly selected for training, and anther $100$ videos are sampled for testing.

\begin{figure*}[h]
	\centering	

	\begin{minipage}[b]{.25\textwidth}
	\centering
		 \includegraphics[width=0.9\linewidth]{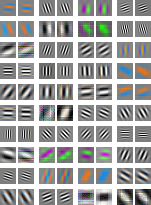}\\
		 (a) Learned units
	\end{minipage}
	\hspace{0mm}
	\begin{minipage}[b]{.25\textwidth}
	\centering
		\includegraphics[width=0.9\linewidth]{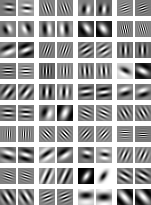}\\
		(b) Fitted Gabors
	\end{minipage}
	\begin{minipage}[b]{.27\textwidth}
	\centering
		 \includegraphics[width=.8\linewidth]{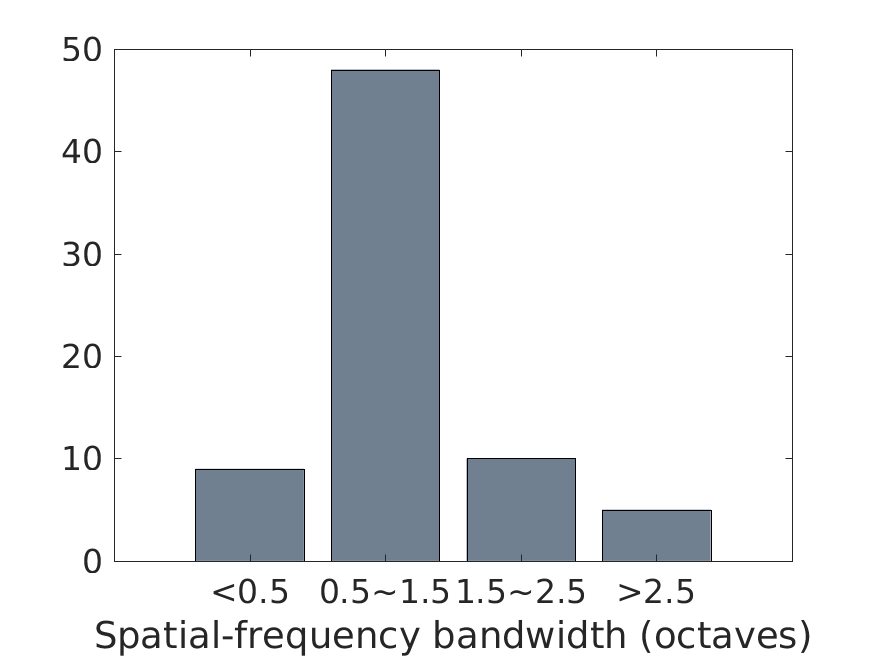} \\
	
	 \includegraphics[width=.8\linewidth]{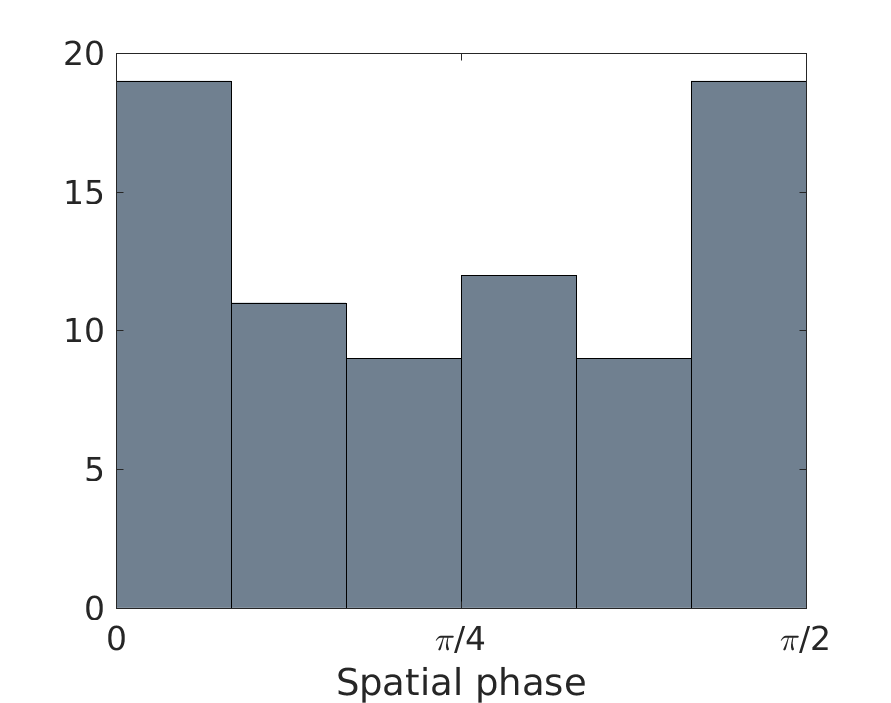} \\
	 (c) Frequency and phase
	\end{minipage}
	 		\caption{\small   Learned results on V1Deform dataset. (a) Learned units. Each block shows two learned units within the same sub-vector. (b) Fitted Gabor patterns. (c) Distributions of spatial-frequency bandwidth (in octaves) and spatial phase $\phi$.  }	
	\label{fig: filters}
\end{figure*}

\subsection{Learned Gabor-like units with quadrature phase relationship}
\label{sect: unit}
In this subsection, we show and analyze the learned units. The size of the filter is $16 \times 16$, with a sub-sampling rate of $8$ pixels. Fig. \ref{fig: filters}(a) displays the learned units, i.e., rows of $W$, on V1Deform dataset. The units are learned with non-parametric $M(\delta)$, i.e., we learn a separate $M(\delta)$ for each displacement and $\delta(x)$ is discretized with an interval of $0.5$. V1-like patterns emerge from the learned units. Moreover, within each sub-vector, the orientations and frequencies of learned units are similar, while the phases are different and approximately follow a quadrature relationship, consistent with the observation of biological V1 simple cells \cite{pollen1981phase,emerson1997quadrature}. Similar patterns can be obtained by using a parametric version of $M(\delta)$. See  Supplementary for more results of learned filters, including filters learned with different dimensions of sub-vectors using different datasets. It is worthwhile to mention that the dimension of sub-vectors is not constrained to be $2$. V1-like patterns also merge when the dimension is $4$ or $6$. 

To further analyze the spatial profiles of the learned units, we fit every unit by a two dimensional Gabor function \cite{jones1987evaluation}: $h(x', y') = A \exp(-(x'/\sqrt{2}\sigma_{x'})^2-(y'/\sqrt{2}\sigma_{y'}))\cos(2\pi fx' + \phi)$, where $(x', y')$ is obtained by translating and rotating the original coordinate system $(x_0, y_0)$: $x' = (x - x_0)\cos\theta+(y-y_0)\sin\theta, y' = -(x - x_0)\sin\theta+(y-y_0)\cos\theta$. The fitted Gabor patterns are shown in Fig. \ref{fig: filters}(b), with the average fitting $r^2$ equal to 0.96 ($\text{std} = 0.04$). The average spatial-frequency bandwidth is 1.13 octaves, with range of $0.12$ to $4.67$. Fig. \ref{fig: filters}(c) shows the distribution of the spatial-frequency bandwidth, where the majority falls within range of $0.5$ to $2.5$. The characteristics are reasonably similar to those of simple-cell receptive fields in the cat \cite{issa2000spatial} (weighted mean $1.32$ octaves, range of $0.5$ to $2.5$) and the macaque monkey \cite{foster1985spatial} (median $1.4$ octaves, range of $0.4$ to $2.6$). To analyze the distribution of the spatial phase $\phi$, we follow the method in \cite{ringach2002spatial} to transform the parameter $\phi$ into an effective range of $0$ to $\pi/2$, and plot the histogram of the transformed $\phi$ in Fig. \ref{fig: filters}(c). The strong bimodal with phases clustering near $0$ and $\pi/2$ is consistent with those of the macaque monkey \cite{ringach2002spatial}.

\begin{figure*}[h]
	\centering	
	\begin{minipage}[b]{.28\textwidth}
	\centering
	
		 \includegraphics[width=0.8\linewidth]{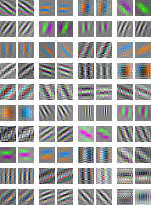}\\
		
		 (a) Learned units
		  \vspace{3mm}
	\end{minipage}
	\begin{minipage}[b]{.6\textwidth}
	\centering
	\begin{minipage}[b]{\textwidth}
	\centering
		 \includegraphics[width=.42\textwidth]{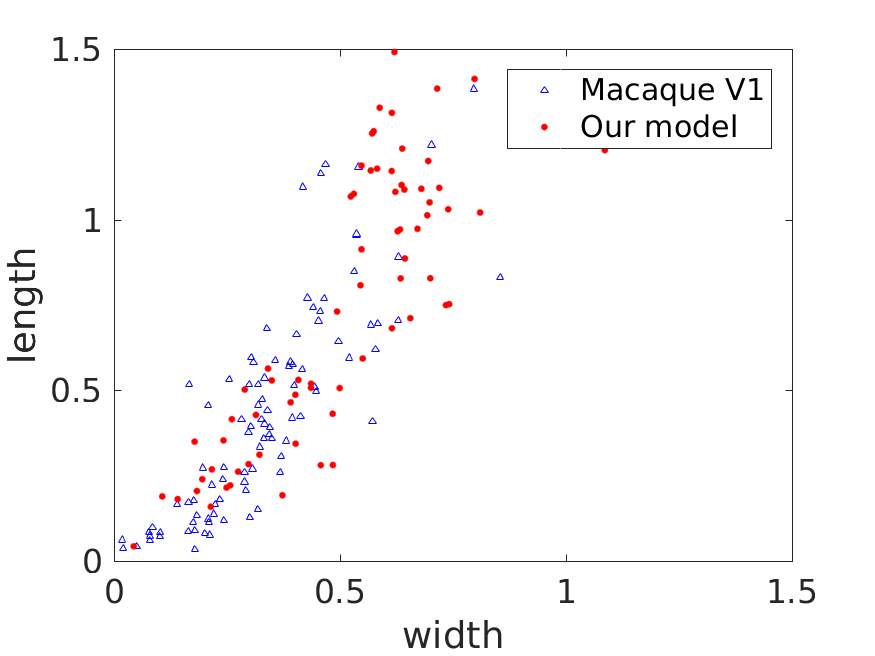} 
	 \includegraphics[width=.42\textwidth]{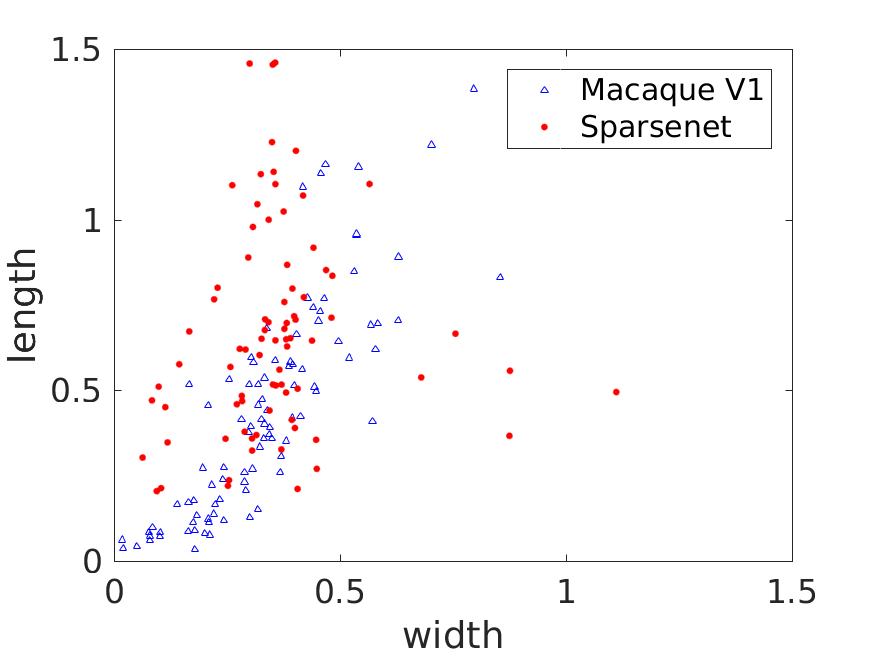}\\ 
	 (b) Gabor envelope shapes of the learned units
	 \end{minipage}\\
	  \begin{minipage}[b]{\textwidth}
	  \centering
	   \includegraphics[width=.28\textwidth]{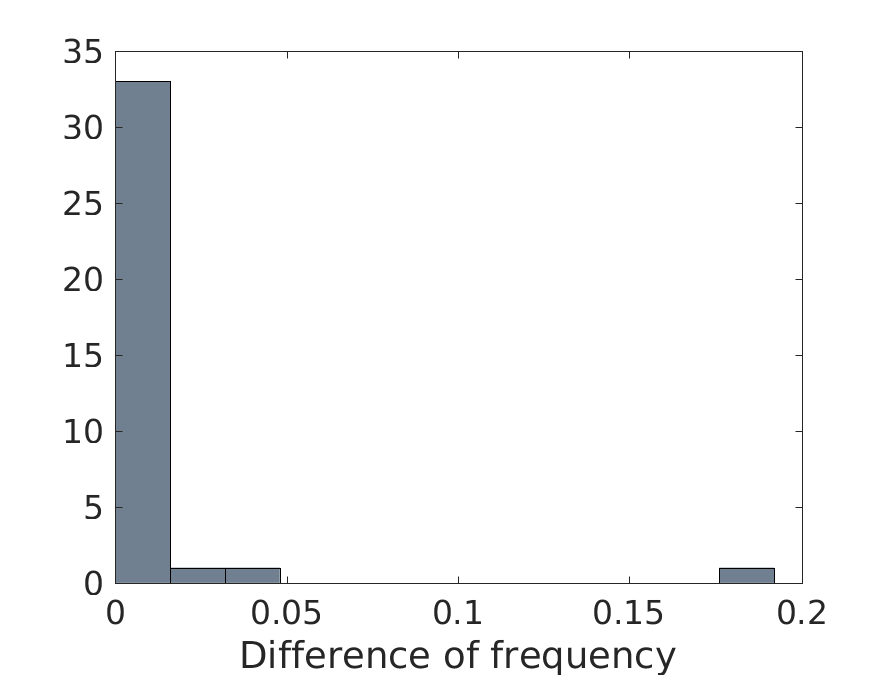}
	   \includegraphics[width=.28\textwidth]{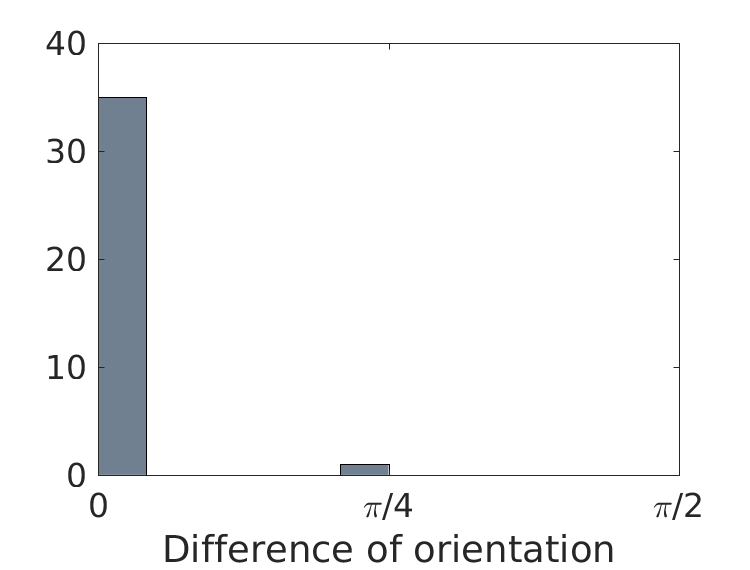}
	   \includegraphics[width=.28\textwidth]{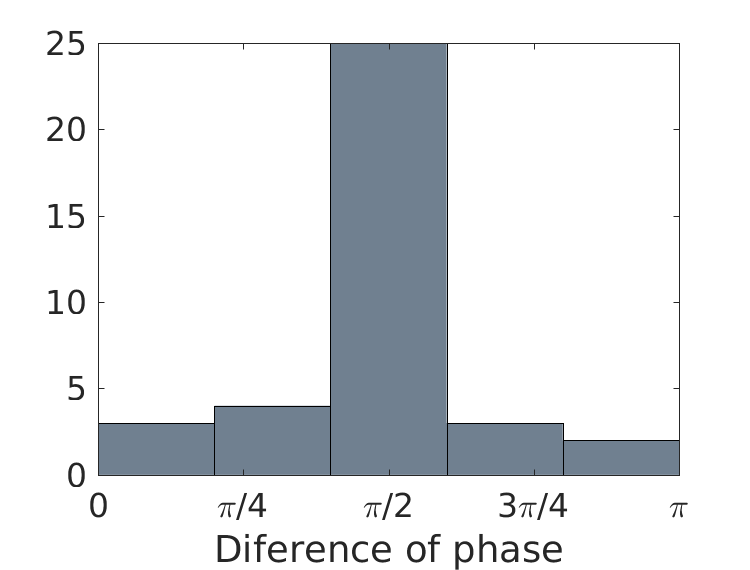}\\
	   (c) Profile of paired units within each sub-vector
	  \end{minipage}
	\end{minipage}
	 		\caption{\small   Learned  results on band-pass image pairs from V1Deform.  (a) Learned units. Each block shows two learned units within the same sub-vector. (b) Distribution of the Gabor envelope shapes in the width and length 2D-plane. (c) Difference of frequency $f$, orientation $\theta$ and phase $\phi$ of paired units within each sub-vector.}	
	\label{fig: filters2}
\end{figure*}

In the above experiment, we fix the size of the convolutional filters ($16 \times 16$ pixels). A more reasonable model is to have different sizes of convolutional filters, with small size filters capturing high-frequency content and large size filters capturing low-frequency content. For fixed-size filters, they should only account for the image content within a frequency band. To this end, we smooth every image by two Gaussian smoothing kernels (kernel size $8$, $\sigma = 1, 4$), and take the difference between the two smoothed images as the input image of the model. The effect of the two smoothing kernels is similar to a bandpass filter so that the input images are constrained within a certain range of frequencies. The learned filters on V1Deform dataset are shown in Fig \ref{fig: filters2}(a). We also fit every unit by a two dimensional Gabor function, resulting in an average fitting $r^2 = 0.83$ ($\text{std}=0.12$). 
Following the analysis of \cite{ringach2002spatial,rehn2007network}, a scatter plot of $n_x = \sigma_xf$ versus $n_y = \sigma_yf$ is constructed in Fig. \ref{fig: filters2}(b) based on the fitted parameters, where $n_x$ and $n_y$ represent the width and length of the Gabor envelopes measured in periods of the cosine waves. Compared to the sparse coding model (a.k.a. Sparsenet) \cite{olshausen1996emergence,olshausen1997sparse}, the units learned by our model have more similar structure to the receptive fields of simples cells of Macaque monkey.

We also quantitatively compare the learned units within each sub-vector in Fig. \ref{fig: filters2}(c). Within each sub-vector, the frequency $f$ and orientation $\theta$ of the paired units tend to be the same. More importantly, most of the paired units differ in phase $\phi$ by approximately $\pi/2$, consistent with the quadrature phase relationship between adjacent simple cells \cite{pollen1981phase,emerson1997quadrature}.  

\subsection{Inference of displacement field}
\label{sect: infer}
We then apply the learned representations to inferring the displacement field $(\delta(x))$ between pairs of frames $(\I_t, \I_{t+1})$. To get valid image patches for inference, we leave out those displacements at image border ($8$ pixels at each side). 

We use non-parametric $M(\delta)$ and the local mixing motion model (Eq. (\ref{eqn: local_mixing})), where the local support $\mathcal{S}$ is in a range of $[-4, +4]$, and $\text{d}x$ is taken with a sub-sampling rate of $2$. After obtaining the inferred displacement field $(\delta(x))$ by the learned model, we also train a CNN model with ResNet blocks \cite{he2016identity} to refine the inferred displacement field. In training this CNN, the input is the inferred displacement field, and the output is the ground truth displacement field, with least-squares regression loss. See  Supplementary for the architecture details of the CNN. For biological interpretation, this refinement CNN is to approximate the processing in visual areas V2-V6 that integrates and refines the motion perception in V1  \cite{gazzaniga2002cognitive,lyon2002evidence,moran1985selective,born2005structure,allman1975dorsomedial}. We learn the models from the training sets of V1Deform and V1FlyingObjects datasets respectively, and test on the corresponding test sets. We also test the model learned from V1FlyingObjects on MPI-Sintel Final, whose frames are resized to $128 \times 128$. 

\begin{figure*}[h]
\begin{center}
	\begin{minipage}{.4\textwidth}
	\centering
\setlength{\tabcolsep}{0.01em}
\begin{tabular}{ccccc}
	$\I_t$ & $\I_{t+1}$ & Gt & FN2 & Ours \\

	\includegraphics[width=.2\linewidth]{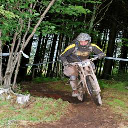} \vspace{0.1cm}& \includegraphics[width=.2\linewidth]{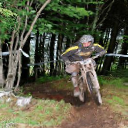} & \includegraphics[width=.2\linewidth]{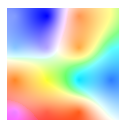} &  \includegraphics[width=.2\linewidth]{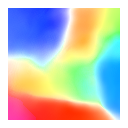} & \includegraphics[width=.2\linewidth]{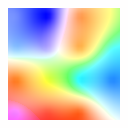}\\
	\includegraphics[width=.2\linewidth]{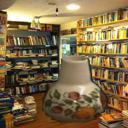} \vspace{0.1cm}& \includegraphics[width=.2\linewidth]{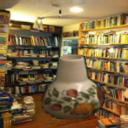} & \includegraphics[width=.2\linewidth]{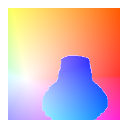} &  \includegraphics[width=.2\linewidth]{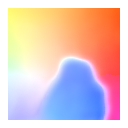} & \includegraphics[width=.2\linewidth]{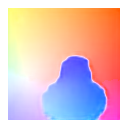} \\
	\includegraphics[width=.2\linewidth]{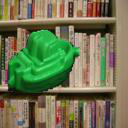} \vspace{0.1cm}& \includegraphics[width=.2\linewidth]{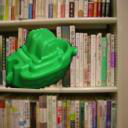} & \includegraphics[width=.2\linewidth]{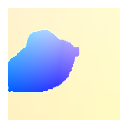} &  \includegraphics[width=.2\linewidth]{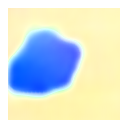} & \includegraphics[width=.2\linewidth]{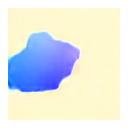} \\
	\includegraphics[width=.2\linewidth]{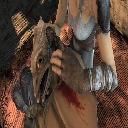} \vspace{0.1cm}& \includegraphics[width=.2\linewidth]{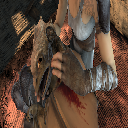} & \includegraphics[width=.2\linewidth]{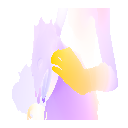} &  \includegraphics[width=.2\linewidth]{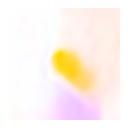} & \includegraphics[width=.2\linewidth]{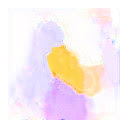}\\
	\includegraphics[width=.2\linewidth]{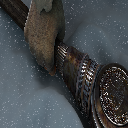} \vspace{0.1cm}& \includegraphics[width=.2\linewidth]{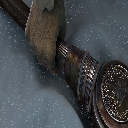} & \includegraphics[width=.2\linewidth]{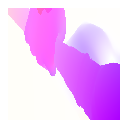} &  \includegraphics[width=.2\linewidth]{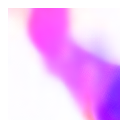} & \includegraphics[width=.2\linewidth]{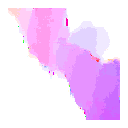}
\end{tabular}
\end{minipage}
\hspace{0.5cm}
\begin{minipage}{.4\textwidth}
	\centering
\setlength{\tabcolsep}{0.01em}
\begin{tabular}{ccccc}
$\I_t$ & $\I_{t+1}$ & Gt & FN2 & Ours \\
\includegraphics[width=.2\linewidth]{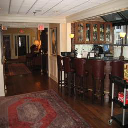} \vspace{0.1cm} & \includegraphics[width=.2\linewidth]{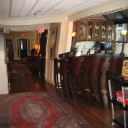} & \includegraphics[width=.2\linewidth]{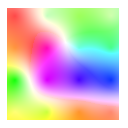}  & \includegraphics[width=.2\linewidth]{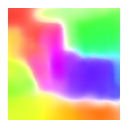} & \includegraphics[width=.2\linewidth]{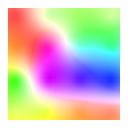}\\
		\includegraphics[width=.2\linewidth]{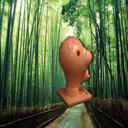} \vspace{0.1cm}& \includegraphics[width=.2\linewidth]{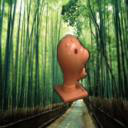} & \includegraphics[width=.2\linewidth]{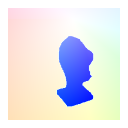} &  \includegraphics[width=.2\linewidth]{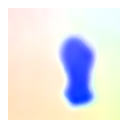} & \includegraphics[width=.2\linewidth]{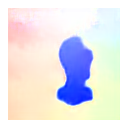} \\
		\includegraphics[width=.2\linewidth]{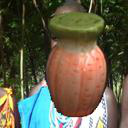} \vspace{0.1cm}& \includegraphics[width=.2\linewidth]{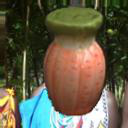} & \includegraphics[width=.2\linewidth]{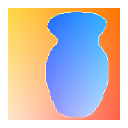} &  \includegraphics[width=.2\linewidth]{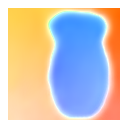} & \includegraphics[width=.2\linewidth]{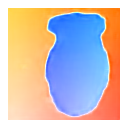} \\ 
		\includegraphics[width=.2\linewidth]{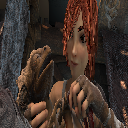} \vspace{0.1cm}& \includegraphics[width=.2\linewidth]{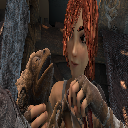} & \includegraphics[width=.2\linewidth]{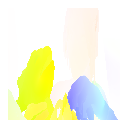} &  \includegraphics[width=.2\linewidth]{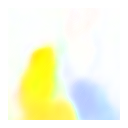} & \includegraphics[width=.2\linewidth]{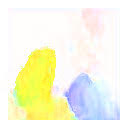}\\
		\includegraphics[width=.2\linewidth]{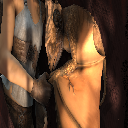} \vspace{0.1cm}& \includegraphics[width=.2\linewidth]{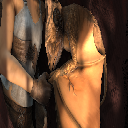} & \includegraphics[width=.2\linewidth]{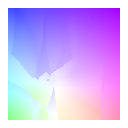} &  \includegraphics[width=.2\linewidth]{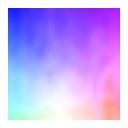} & \includegraphics[width=.2\linewidth]{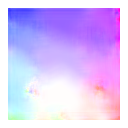}
\end{tabular}
\end{minipage}
\end{center}
\caption{\small   Examples of inferred displacement field on V1Deform, V1FlyingObjects and MPI-Sintel. For each block, from left to right are $\I_t$, $\I_{t+1}$, ground truth displacement field and inferred displacement field by FlowNet 2.0 model and our learned model respectively. For each dataset, we show the result of FlowNet 2.0 model with lower AEE between the pre-trained and trained ones. The displacement fields are color coded \cite{liu2010sift}. See  Supplementary for the color code.} 
\label{fig: infer_non_para1}
\end{figure*}

Table \ref{table: infer} summarizes the average endpoint error (AEE) of the inferred results. We compare with several baseline methods, including FlowNet 2.0 and its variants \cite{DFIB15,IMKDB17}. For baseline methods, we test the performance using two models: one retrained on our datasets (`trained') and the released model by the original authors which are pre-trained on large-scale datasets (`pre-trained'). Note that for MPI-Sintel, a pre-trained baseline model gives better performance compared to the one trained on V1FlyingObjects, probably because these methods train deep and complicated neural networks with large amount of parameters to predict optical flows in supervised manners, which may require large scale data to fit and transfer to different domains. On the other hand, our model can be treated as a simple one-layer auto-encoder network, accompanied by weight matrices representing motions. As shown in Table \ref{table: infer}, our model has about $88$ times fewer parameters than FlowNet 2.0 and $21$ times fewer parameters than the light FlowNet2-C model. We achieve competitive performance compared to these baseline methods. Fig. \ref{fig: infer_non_para1} displays several examples of the inferred displacement field.  Inferred results from the FlowNet 2.0 models are shown as a qualitative comparison. For each dataset, we show the result of FlowNet 2.0 model with lower AEE between the pre-trained and trained ones. 
%

\begin{table*}[h]
\caption{\small   Average endpoint error of the inferred displacement and number of parameters. Abbreviation FN2 refers to FlowNet 2.0. `Ours-ref' indicates that the results are post-processed by the refinement CNN.} 
\label{tab: infer_comp}
\centering
 \small
\begin{tabular}{lccccccc}
\toprule
 & \multicolumn{2}{c}{V1Deform}            & \multicolumn{2}{c}{V1FlyingObjects}  &  \multicolumn{2}{c}{MPI-Sintel}& \multirow{2}{*}{\# params (M)}     \\ 
 & pre-trained & trained & pre-trained & trained  & pre-trained & trained & \\ \midrule
 FN2-C & 1.324 & 1.130  & 0.852 & 1.034 & 0.363 & 0.524 & 39.18 \\
 FN2-S & 1.316 & 0.213 & 0.865 & 0.261 & 0.410 & 0.422 & 38.68 \\
 FN2-CS & 0.713 & 0.264 & 0.362 & 0.243 & 0.266 & 0.346 & 77.87 \\
 FN2-CSS & 0.629 & 0.301 & 0.299 & 0.303  & 0.234 & 0.450 & 116.57 \\
 FN2 & 0.686 & 0.205  & 0.285 & 0.265 & 0.146 & 0.278 & 162.52 \\
 Ours & - & 0.258 & - & 0.442 & - & 0.337 & 1.82 \\
 Ours-ref & - & {\bf 0.156} & - & {\bf 0.202} & - & {\bf 0.140} & 1.84\\

 \bottomrule
\end{tabular}
\label{table: infer}
\end{table*}

\subsection{Unsupervised learning}
\label{sect: unsupervised}
\begin{figure*}[h]
\begin{center}
	\begin{minipage}{.42\textwidth}
	\centering
	\includegraphics[width=.96\linewidth]{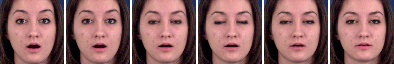}\\
	\includegraphics[width=.96\linewidth]{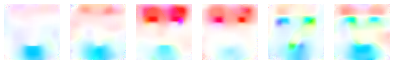}\\
\setlength{\tabcolsep}{0.01em}
\end{minipage}
	\begin{minipage}{.42\textwidth}
	\centering
\includegraphics[width=.94\linewidth]{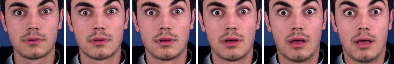}\\
 \includegraphics[width=.94\linewidth]{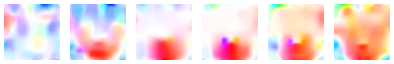}\\
\setlength{\tabcolsep}{0.01em}
\end{minipage}
\end{center}

\caption{\small   Examples of inferred displacement fields by unsupervised learning on MUG Facial Expression dataset. Within each block, the top row shows the observed image frames, while the bottom row shows the inferred color-coded displacement fields \cite{liu2010sift}. See  Supplementary for the color code.
}
\label{fig: infer_unsupervised}
\end{figure*}

We further perform unsupervised learning of the proposed model, i.e., without knowing the displacement field of training pairs. For unsupervised learning, we scale the images to size $64 \times 64$. The size of the filters is $8 \times 8$, with a sub-sampling rate of $4$ pixels. Displacements at the image border ($4$ pixels at each side) are left out. We train the model on MUG Facial Expression and V1FlyingObjects datasets. Fig. \ref{fig: infer_unsupervised} shows some examples of inferred displacement fields on the testing set of MUG Facial Expression. The inference results are reasonable, which capture the motions around eyes, eyebrows, chin, or mouth. For the model trained on V1FlyingObjects, we test on the testing set of V1FlyingObjects and MPI-Sintel. Table \ref{table: infer_unsup} summarizes the quantitative results. We include comparisons with several baseline methods for unsupervised optical flow estimation: Unsup \cite{jason2016back} and UnFlow \cite{meister2018unflow} and its variants, which are also trained on V1FlyingObjects. The proposed model achieves better performance compared to baseline methods. See  Supplementary for qualitative comparisons and more inference results. 

\begin{table*}[h]
\caption{\small   Average endpoint error of the inferred displacement in unsupervised learning.}
\label{tab: infer_comp}
\centering
 \small
\begin{tabular}{lccccccc}
\toprule
 & Unsup &  UnFlow-C & UnFlow-CS & UnFlow-CSS  & Ours     \\ \midrule
V1FlyingObjects (train)   & 0.379  & 0.336  & 0.374& 0.347 & {\bf 0.245}  \\ 
V1FlyingObjects (test)   & 0.811  & 0.399  & 0.394 & 0.453 & {\bf 0.316} \\ 
MPI-Sintel    & 0.440  & 0.198  & 0.248 & 0.202 & {\bf 0.101} \\
 \bottomrule
\end{tabular}
\label{table: infer_unsup}
\end{table*}

\begin{table}[h]
\caption{\small   Ablation study measured by average endpoint error (AEE) of motion inference and mean squared error (MSE) of multi-step animation, learned from V1Deform dataset.}
\label{tab: ablation}
\begin{center}
\begin{minipage}{.45\textwidth}
\centering
 \small
\begin{tabular}{lccccc}
\toprule
& \multicolumn{5}{c}{Sub-vector dimension}\\
& 2 & 4 & 6 & 8 & 12 \\ \midrule
AEE & 0.258 & 0.246 & 0.238 & 0.241 & 0.0.246\\
MSE & 8.122 & 7.986 & 7.125 & 7. 586 & 8.017 \\
 \bottomrule
\end{tabular}
\end{minipage} \\
\vspace{1mm}
\begin{minipage}{.45\textwidth}
\centering
 \small
\begin{tabular}{lccc}
\toprule
&  \multicolumn{3}{c}{Sub-sampling rate} \\
 & 4 & 8 & 16 \\ \midrule
AEE & 0.383 & 0.258 & 0.312 \\
MSE &  11.139 & 8.122 & 10.293 \\
 \bottomrule
\end{tabular}
\end{minipage}
\end{center}
\end{table}

\subsection{Multi-step frame animation}
\label{supp: frame_animation}
\begin{figure}[h]
\centering	
\rotatebox{90}{\hspace{4mm}{\footnotesize gt}}	\includegraphics[height=.14\linewidth]{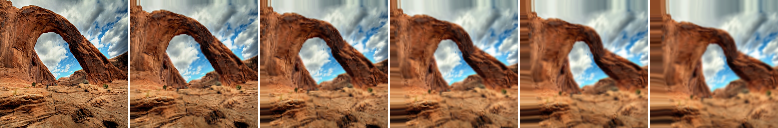} \\
\rotatebox{90}{\hspace{4mm}{\footnotesize syn}} 
\includegraphics[height=.14\linewidth]{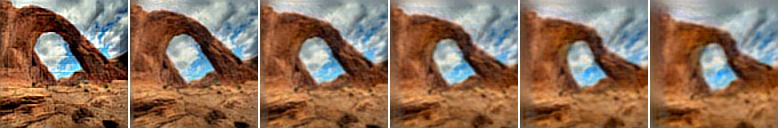}
 \\ \vspace{1mm}
\caption{\small   Example of multi-step animation. For each block, the first row shows the ground truth frame sequences, while the second row shows the animated frame sequences.}
	\label{fig: predict}
\end{figure}

The learned model is also capable of multi-step frame animation. Specifically, given the starting frame $\I_0(x)$ and a sequence of displacement fields $\{\delta_1(x), ..., \delta_T(x), \forall x \}$, we can animate the subsequent multiple frames $\{\I_1(x), ..., \I_T(x)\}$ using the learned model. We use the model with local mixing. We introduce a re-encoding process when performing multi-step animation. At time $t$, after we get the next animated frame $\I_{t+1}$, we take it as the observed frame at time $t+1$, and re-encode it to obtain the latent vector $v_{t+1}$ at time $t+1$. 
Fig. \ref{fig: predict} displays two examples of animation for 6 steps, learned with non-parametric version of $M$ on V1Deform and V1FlyingObjects. The animated frames match the ground truth frames well. See Supplementary for more results.

\subsection{Frame interpolation}
\label{supp: interpolation}
Inspired by the animation and inference results, we show that our model can also perform frame interpolation, by combining the animation and inference together. Specifically, given a pair of starting frame $\I_0$ and end frame $\I_T$, we want to derive a sequence of frames $(\I_0, \I_1, ..., \I_{T-1}, \I_{T})$ that changes smoothly. Let $v_0(x) = W\I_0[x]$ and $v_T(x) = W\I_T[x]$ for each $x \in D$. At time step $t+1$, like the inference, we can infer displacement field $\delta_{t+1}(x)$ by steepest descent:
\begin{align}
	\hat{v}_{t+1}(x, \delta) &= \sum_{\text{d} x \in \mathcal{S}} \nolimits M(\delta, \text{d} x) {v}_{t}(x + \text{d} x), \\ 
	\delta_{t+1}(x) &= \arg\min_{\delta \in \Delta}\sum_{k=1}^K \left\|v_T^{(k)} - \hat{v}^{(k)}_{t+1}(x, \delta)\right\|^2. 
\end{align}
Like the animation, we get the animated frame $\I_{t+1}$ by decoding $\hat{v}_{t+1}(x, \delta_{t+1}(x))$, and then re-encode it to obtain the latent vector $v_{t+1}(x)$.

\begin{figure}[h] 
	\centering
\includegraphics[height=.14\linewidth]{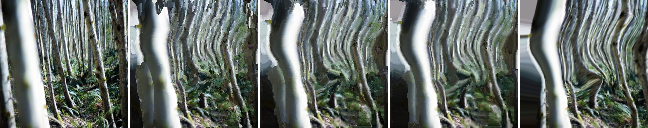}\\ \vspace{1mm} \includegraphics[height=.14\linewidth]{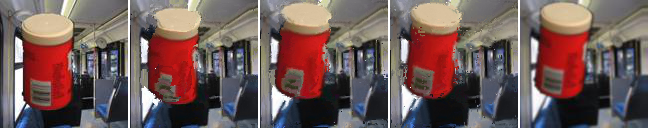}
\caption{\small   Examples of frame interpolation, learned with non-parametric $M$. For each block, the first frame and last frame are given, while the frames between them are interpolated frames.}
\label{fig: interpolation}
\end{figure}

The algorithm stops when $\I_t$ is close enough to $\I_T$ (mean pixel error $<10$). Fig. \ref{fig: interpolation} shows four examples, learned with non-parametric $M$ on V1Deform and V1FlyingObjects. For $96.0\%$ of the testing pairs, the algorithm can accomplish the frame interpolation within $10$ steps.  With this algorithm, we are also able to infer displacements larger than the acceptable range of $\delta$ by accumulating the displacements along the interpolation steps. See Supplementary for more results.

\subsection{Ablation study}
We perform ablation studies to analyze the effect of two components of the proposed model: (1) dimensionality of sub-vectors and (2) sub-sampling rate. Besides comparing the average endpoint error (AEE) of motion inference, we also test if the learned model can make accurate multi-step animation of image frames given the sequence of displacement fields. Per pixel mean squared error (MSE) of the predicted next five image frames is reported. Table \ref{tab: ablation} summarizes the results learned from V1Deform dataset. The dimensionality of sub-vectors controls the complexity of the motion matrices, which is set to a minimum of $2$ in the experiments. As the dimensionality of sub-vectors increases, the error rates of the two tasks decrease first and then increase. On the other hand, sub-sampling rate can be changed to make the adjacent image patches connect with each other more loosely or tightly. As shown in Table \ref{tab: ablation}, a sub-sampling rate of $8$, which is half of the filter size, leads to the optimal performance. 

\section{Conclusion} 

This paper proposes a simple representational model that couples vector representations of local image contents and matrix representations of local motions, so that the vector representations are equivariant. 
Unlike existing models for V1 that focus on statistical properties of natural images or videos, our model serves a direct purpose of perception of local motions. 
Our model learns Gabor-like units with quadrature phases. We also give biological interpretations of the learned model and connect it to the spatiotemporal energy model. It is our hope that our model adds to our understanding of motion perception in V1. 

Our motion model can be integrated with the sparse coding model. For sparse coding, we can keep the top-down decoder of the tight frame auto-encoder for each image frame, and impose sparsity on the number of sub-vectors that are active. For motion, we then assume that each sub-vectors is transformed by a sub-matrix that represents the local displacement. 

This paper assumes linear decoder for image frames and matrix representation of local motion. We can generalize the linear decoder to a neural network and generalize the matrix representation to non-linear transformation modeled by non-linear recurrent network. 

In our future work, we shall study the inference of ego-motion, object motions and 3D depth information by generalizing our model based on equivariant vector representations and their transformations. We shall also apply our model to stereo in binocular vision.  


\section*{Acknowledgement}
The work is supported by NSF DMS-2015577, ONR MURI project N00014-16-1-2007, DARPA XAI project N66001-17-2-4029, and XSEDE grant ASC170063. We thank Prof. Tai Sing Lee for sharing his  knowledge and insights on V1. 

\clearpage

%
%

\bibliographystyle{aaai22}
\bibliography{aaai22}
\end{document}